\documentclass[acmtog]{acmart}

\usepackage[ruled]{algorithm2e} %

\SetAlFnt{\small}
\SetAlCapFnt{\small}
\SetAlCapNameFnt{\small}
\SetAlCapHSkip{0pt}

\copyrightyear{2026}
\acmYear{2026}
\setcopyright{cc}
\setcctype{by}
\acmConference[SA Conference Papers '26]{SIGGRAPH Asia 2026 Conference Papers}{December 01--04, 2026}{Kuala Lumpur, Malaysia}
\acmBooktitle{SIGGRAPH Asia 2026 Conference Papers (SA Conference Papers '26), December 01--04, 2026, Kuala Lumpur, Malaysia}
\acmDOI{10.1145/3829340.3842234}
\acmISBN{979-8-4007-2842-6/2026/12}

\usepackage{cleveref}
\usepackage{siunitx}
\usepackage{colortbl} %

\graphicspath{{figures/}}

\newcommand{\chosenrow}{\rowcolor{black!8}}

\newcommand{\vs}{\vec{s}}
\newcommand{\vn}{\vec{n}}
\newcommand{\hz}{\hat{z}}
\newcommand{\dw}{d\omega}
\newcommand{\Ls}{L(\vs)} %
\newcommand{\Ln}{L(\vn)} %
\newcommand{\En}{E(\vn)} %

\newcommand{\kn}{k(\vn)}
\newcommand{\kl}{k_\ell}
\newcommand{\Ylm}{Y_{\ell m}}
\newcommand{\Yn}{Y_{\ell m}(\vn)}
\newcommand{\Elm}{E_{\ell m}}
\newcommand{\Llm}{L_{\ell m}}

\begin{document}

\title{An Eternal Irradiance Camera}

\author{Jeremy Klotz}
\orcid{0009-0005-9408-3541}
\affiliation{%
\institution{Computer Science Department, Columbia University}
\city{New York}
\state{NY}
\country{USA}}
\email{jklotz@cs.columbia.edu}

\author{Shree K. Nayar}
\orcid{0000-0002-6452-6998}
\affiliation{%
\institution{Computer Science Department, Columbia University}
\city{New York}
\state{NY}
\country{USA}}
\email{nayar@cs.columbia.edu}

\begin{teaserfigure}
    \centering
    \includegraphics{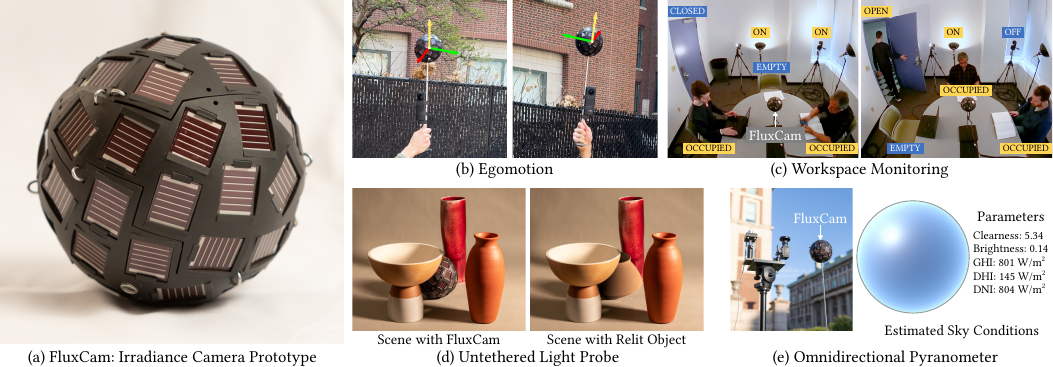}
    \vspace{-0.1in}
    \caption{
        \textbf{An irradiance camera and its applications.}
        (a)~We present a novel camera that measures the total illumination (irradiance) falling on every point of a sphere. Our prototype camera, called FluxCam, has a minimalist design: it uses a very small number of detectors to both measure irradiance and harvest energy. This allows FluxCam to be completely powered by the light falling on it, i.e.,~it can read out and wirelessly transmit its measurements at a framerate of 30 FPS or more without a battery, cable, or external power supply. FluxCam can be used for (b)~estimating egomotion and (c)~monitoring a workspace. In these cases, the camera is not only self-powered and untethered but also privacy preserving since it does not capture the visual details needed to identify individuals. FluxCam can also be used as (d)~an untethered light probe for diffuse relighting and (e)~an omnidirectional pyranometer for analyzing the sky's illumination.
    }
    \label{fig:teaser}
\end{teaserfigure}

\begin{abstract}
A conventional camera uses millions of pixels to measure radiance from all directions within its field of view. We present an omnidirectional \textit{irradiance camera} that measures the irradiance function---the illumination incident upon every point on a sphere. The irradiance function varies smoothly over the sphere and hence is bandlimited. We have analyzed this function in the frequency domain and have shown that it is well approximated by a weighted sum of the first seven degrees of spherical harmonics. As a result, the irradiance function can be accurately reconstructed from a small number of samples. This implies that an irradiance camera does not need millions of detectors (pixels)---just a handful of measurements suffice. This brings two major benefits. First, the camera consumes such little energy that it can be completely powered by the light falling on its detectors. Second, it does not capture the visual details needed to identify an individual, and hence privacy is preserved. We have built a prototype irradiance camera, called FluxCam, using 49 detectors arranged on the surface of a sphere. In a well-lit indoor environment, FluxCam can read out and wirelessly transmit its measurements at 30 frames per second using energy harvested from the light falling on it (i.e.,~without a battery, cable, or external power supply). We show how FluxCam can be used as an optical gyroscope for computing rotation, to monitor a workspace, as an untethered light probe for diffuse relighting, and as an omnidirectional pyranometer for estimating sky conditions and determining the best orientation of a solar panel.

\end{abstract}



\maketitle

\section{What is an Irradiance Camera?}

\begin{figure*}[t]
    \centering
    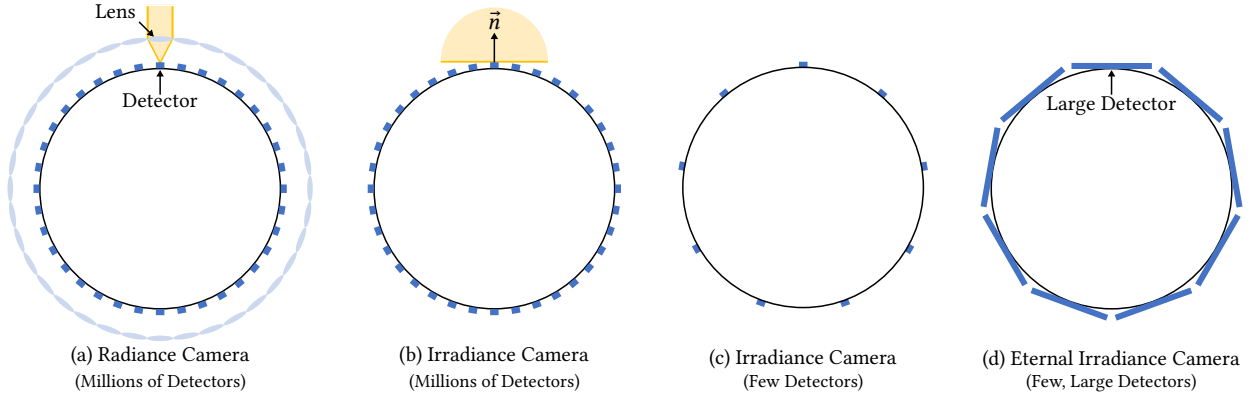
    \vspace{-0.1in}
    \caption{
        \textbf{What is an irradiance camera?}
        (a)~A conventional omnidirectional camera uses millions of pixels to measure radiance from every direction. Each pixel can be modeled by a detector that views the scene through a lens.
        (b)~If we remove the lenses, each detector measures irradiance received from a hemisphere. In effect, this camera measures the illumination falling on every point of a sphere. Since it measures the irradiance function, we call it an irradiance camera.
        (c)~The irradiance function in any environment is extremely smooth, and hence an irradiance camera only needs a handful of detectors to measure it.
        (d)~Since so few detectors are required, we can use large detectors that cover the entire surface of the sphere. Large detectors not only produce measurements with a wide dynamic range, but can also be used as solar cells that harvest energy from the incident illumination. This harvested energy can be used to fully power the camera using the light falling on it. As a result, the camera can operate forever, as an eternal camera, without a battery, cable, or external power supply.
    }
    \label{fig:what-is}
    \vspace{-0.1in}
\end{figure*}

Consider a scene viewed from a point. The brightness of the scene can be represented as $\Ln$, where $L$ is the scene radiance in the direction $\vn$. 
The goal of a conventional omnidirectional camera is to measure the radiance function $\Ln$. One way to represent such a camera is by arranging detectors (pixels) on a sphere, with a lens above each detector.\footnote{This simplified architecture can be used to represent a wide class of camera designs: a camera with a single lens, a spherical camera with two lenses, a catadioptric camera with a lens and mirror, or a light probe consisting of a camera and mirror ball.} This architecture is illustrated in \cref{fig:what-is}(a), where each detector measures radiance from a single direction. Since a conventional camera densely samples the radiance function, it can be thought of as a \textit{radiance camera}.

Suppose now that we remove the lenses above the detectors in \cref{fig:what-is}(a), as shown in \cref{fig:what-is}(b). Each detector now measures the total irradiance it receives from a hemisphere, which can be written as an integral over the radiance function,
\begin{equation}
    \En = \int_{\vs \in \mathbb{S}^2} \Ls \, \max (\vn \cdot \vs, 0) \, \dw, \label{eq:En}
\end{equation}
where $\vn$ is the vector normal to the detector, $\dw$ is the differential solid angle, and the max operator accounts for the fact that a detector only receives light from a hemisphere. 
\Cref{eq:En} defines the irradiance function, which is fundamental to our work. Given that the detectors in \cref{fig:what-is}(b) sample the irradiance function, we refer to this camera as an \textit{irradiance camera}.

The irradiance function at any point in a scene has a simple visual interpretation: it is equivalent (up to a scale factor) to the brightness distribution of an infinitesimally small Lambertian sphere with constant albedo placed at the point. Since each point on the sphere receives light from an entire hemisphere, the irradiance function is smooth, regardless of the complexity of the scene. It does not include the fine-grained details in the corresponding radiance function.

Although it is smooth, we posit that the irradiance function is an expressive signal for a wide class of visual processing tasks. In fact, the irradiance function at every point in space is a complete representation of Gershun's light field~\cite{gershunLightField1939}, which describes the flow of illumination. When an irradiance camera moves through space, its irradiance function encodes its motion. 
Furthermore, there exists a large class of vision tasks, called lightweight tasks, that can be solved without sensing radiance functions
\cite{klotzMinimalistVisionFreeform2025}. An irradiance camera can be used to solve such tasks in an efficient manner.

\begin{figure}[t]
    \centering
    \includegraphics{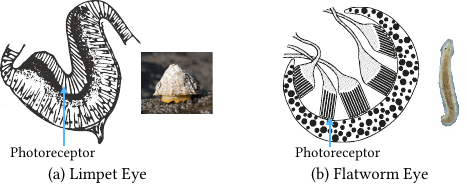}
    \vspace{-0.1in}
    \caption{\textbf{Irradiance cameras in nature.} The eyes of (a)~a limpet and (b)~a flatworm resemble an irradiance camera. These eyes do not have lenses. Instead, the photoreceptors (detectors) are arranged on a curved surface. 
    Even though these eyes do not sense detailed images, they provide enough visual information for the organisms to function in their environments. Figures adapted from \citet{landAnimalEyes2002}.}
    \label{fig:nature}
    \vspace{-0.2in}
\end{figure}

In nature, several organisms are able to function in their worlds by sensing just the irradiance function~\cite{landAnimalEyes2002}. \Cref{fig:nature} shows the eye of a limpet and the eye of a flatworm, neither of which has a lens. In each case, the photoreceptors (detectors) are densely packed on a curved surface, similar to the detectors of the irradiance camera shown in \cref{fig:what-is}(b). Each photoreceptor measures irradiance due to a different field of view. Naturally, these eyes do not form a detailed image; however, they provide the organisms with sufficient visual information to navigate around and survive in their environment.

It has been shown analytically that the irradiance function is smooth (i.e.,~bandlimited), regardless of the complexity of the radiance function~\cite{basriLambertianReflectanceLinear2003,ramamoorthiRelationshipRadianceIrradiance2001}. Consequently, an irradiance camera does not need millions of detectors (as in \cref{fig:what-is}(b)) to measure the irradiance function---just a handful of detectors suffice (\cref{fig:what-is}(c)). Since so few detectors are needed, an irradiance camera can use large detectors that cover the entire surface of the sphere (\cref{fig:what-is}(d)). These large detectors collect more light, allowing them to not only produce high dynamic range measurements but also harvest enough energy to power the camera.

We have built the prototype irradiance camera shown in \cref{fig:teaser}(a) using 49 detectors arranged on the surface of a sphere. 
We refer to this prototype as FluxCam since its measurements are proportional to the light flux received by the surface of the sphere.
Each detector is a photovoltaic cell that measures irradiance over a wide dynamic range, spanning $2\,\unit{\lux}$ to $13{,}000\,\unit{\lux}$. When the photovoltaics are not used for sensing, they harvest energy from the light falling on them to power the camera. In a well-lit indoor environment,
FluxCam can read out and wirelessly transmit its measurements at 30 frames per second (FPS). In a bright outdoor environment, it can function at up to 200 FPS. So long as there is sufficient light to power the camera, it can operate forever, without a battery or power supply. For this reason, we refer to it as an \textit{eternal irradiance camera}.

From the 49 measurements made by the camera, we reconstruct the irradiance function. We have validated the accuracy of our reconstruction method using both simulated and real-world measurements, in indoor and outdoor settings. 
We have used FluxCam for computing egomotion (\cref{fig:teaser}(b)) and monitoring a workspace by determining which desks are occupied, when a door is open, and which lamps are activated (\cref{fig:teaser}(c)). In these cases, the irradiance camera is not only self-powered and untethered but also privacy preserving since it does not capture the fine-grained visual details needed to identify individuals. In addition, we have used FluxCam as an untethered light probe for diffuse relighting (\cref{fig:teaser}(d)) and as an omnidirectional pyranometer for estimating the current sky conditions and determining the best orientation of a solar panel to maximize its harvested energy (\cref{fig:teaser}(e)).

\vspace{-0.15in}
\section{Related Work}

Prior work has established that the irradiance function in any environment is bandlimited~\cite{basriLambertianReflectanceLinear2003,ramamoorthiRelationshipRadianceIrradiance2001,ramamoorthiSignalProcessingFrameworkInverse2001}. Building on this result, we show that the irradiance function can be accurately reconstructed using a very small number of samples (measurements). This enables us to build a camera that consumes very little energy and hence is fully self-powered.

A popular approach for measuring the radiance function is to use a light probe, consisting of a mirror and a camera that observes it from a distance~\cite{debevecRenderingSyntheticObjects1998}. Subsequent works have extended the mirror probe for measuring high dynamic range (HDR) radiance functions either by adding a diffuse probe~\cite{reinhardHighDynamicRange2010} or by attaching diffuse patches to an existing mirror probe~\cite{debevecSingleshotLightProbe2012}. %
All of these devices capture a detailed representation of incident light, which can be used for rendering objects with arbitrary shapes and material properties. However, the mirror and diffuse probes must be fully visible to the camera and hence cannot be placed at locations at which they are  entirely, or even partially, occluded from the camera. Our irradiance camera, on the other hand, is an untethered light probe that measures the complete irradiance function for any location it can be physically placed at.

Although the irradiance function is not as expressive as the radiance function, it has proved useful in real-time rendering pipelines.  \citet{wardRayTracingSolution1988} proposed irradiance caching as a technique for speeding up the computation of global illumination. 
\citet{gregerIrradianceVolume1998} introduced the irradiance volume to represent the irradiance function at every point in space.
\citet{ramamoorthiEfficientRepresentationIrradiance2001} proposed a compact representation of the irradiance function, along with a fast algorithm for rendering diffuse objects. 
Our irradiance camera is designed to directly measure the irradiance function in real environments.

Prior work has shown that low-dimensional visual signals are sufficient for solving a variety of vision tasks, including human activity recognition~\cite{bobickRecognitionHumanMovement2001a,daiPrivacyPreservingActivityRecognition2015}, person detection~\cite{nakashimaDevelopmentPrivacypreservingSensor2010a}, and even image classification~\cite{torralba80MillionTiny2008}. 
In this context, we show that the measurements produced by an irradiance camera can be used for monitoring a workspace, i.e.,~determining which desks are occupied, when a door is open, and which lamps are on. We are able to solve these tasks with just $49$ measurements, in part, because our detectors have high sensitivity and wide dynamic range. Interestingly, 
since the camera does not capture fine-grained visual details, it is difficult to identify an individual from the measurements, which is beneficial in applications where privacy needs to be preserved.

\begin{figure*}[t]
    \centering
    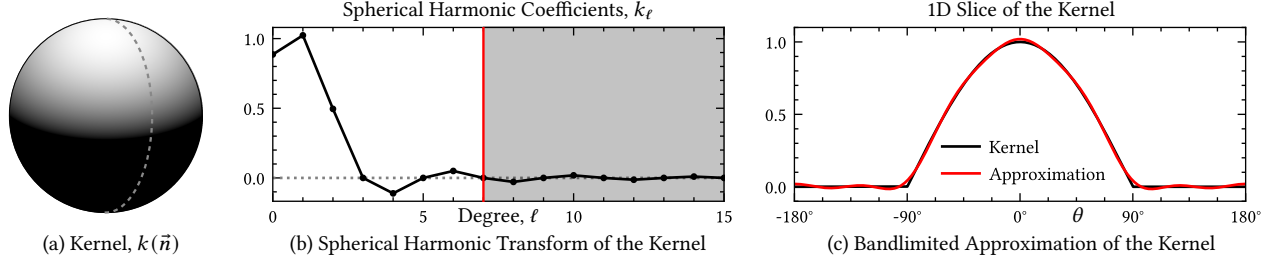
    \vspace{-0.1in}
    \caption{
        \textbf{Spherical harmonic analysis of the kernel.}
        (a)~The irradiance function is the result of convolving the radiance function with the kernel $\kn$, which is defined as the cosine of the zenith angle ($\theta$), but clamped to be non-negative.
        (b)~The spherical harmonic coefficients of the kernel as a function of the spherical harmonic degree. The coefficients $\kl$ tend toward 0 as the degree (frequency) increases, as a result of the kernel being a low-pass filter. We see from the plot that $\kl$ is approximately zero for any degree $\ell \ge 7$ (shaded gray region to the right of the vertical red line). Therefore, the kernel can be approximated by a weighted sum of the first seven spherical harmonics.
        (c)~A 1D slice of the kernel through the gray dotted line in~(a). The bandlimited approximation computed from the first seven spherical harmonics (red line) closely follows the kernel (black line).
    }
    \label{fig:kernel}
    \vspace{-0.1in}
\end{figure*}

In meteorology and solar energy, pyranometers are widely used for obtaining the irradiance measurement corresponding to a single direction~\cite{duffieSolarEngineeringThermal2006,augustineSURFRADNationalSurface2000}. These measurements, made over time, can be used to obtain coarse estimates of the condition of the sky~\cite{longIdentificationClearSkies2000,duchonEstimatingCloudType1999}. Others have used multiple pyranometers in different, fixed orientations to obtain more detailed descriptions of the sky condition~\cite{lammNewMethodDetermination1987,leeMultiDirectionalPyranometerCUBEi2025,faimanMultipyranometerInstrumentObtaining1992,baltazarImprovedMethodologyEvaluate2015}. 
The irradiance camera is essentially an omnidirectional pyranometer that provides these measurements  without any moving parts. We use the camera to demonstrate multiple applications in this context: estimating the sky conditions, decomposing irradiance into sun and sky contributions, and finding the best orientation of a solar panel.

Our work is related to previous work on self-powered light sensing and imaging. A variety of light sensors (pixels) have been proposed that use the incident illumination to power the measurement and wireless transmission~\cite{nayarCricketSelfPoweredChirping2024,zhangOptoSenseUbiquitousSelfPowered2020,zhangSozuSelfPoweredRadio2019}. \Citet{nayarSelfPoweredCameras2015} developed a $30\times40\,\unit{\text{px}^2}$ camera that is powered by the light falling on the pixels. This camera harvested enough energy to read out the image, but not enough to wirelessly transmit it. The minimalist camera~\cite{klotzMinimalistVisionFreeform2025} used a handful of freeform pixels whose shapes are automatically learned for a given task. Since the power consumption of any camera is roughly linear in the number of detectors, the minimalist camera was able to read out and wirelessly transmit using energy harvested from the incident light. Given that the irradiance camera uses very few detectors, it is also powered by the light falling on it. However, rather than designing the camera for a specific task, our goal is to develop a general-purpose device that accurately measures the irradiance function for a variety of downstream tasks.

\vspace{-0.05in}
\section{Measuring the Irradiance Function}
We begin by analyzing the properties of the irradiance function using spherical harmonics.

\vspace{-0.05in}
\subsection{Degrees of Freedom}
Prior work by \citet{basriLambertianReflectanceLinear2003} and \citet{ramamoorthiSignalProcessingFrameworkInverse2001} has shown that the irradiance function (\cref{eq:En}) can be written as the spherical convolution\footnote{We refer the interested reader to \citet{driscollComputingFourierTransforms1994} for an introduction to convolution on the sphere.} of the radiance function $\Ln$ with a kernel given by
\begin{equation}
    \kn = \max (\vn \cdot \hz, 0), \label{eq:k}
\end{equation}
where $\hz$ is the zenith vector. \Cref{fig:kernel}(a) shows the kernel, which is a truncated cosine function. Since the kernel is extremely smooth and broad, it acts as a low-pass filter on the radiance function. Building on this prior work, we analyze the kernel's bandwidth. \Cref{fig:kernel}(b) plots the spherical harmonic coefficients $\kl$ of the kernel as a function of the spherical harmonic degree $\ell$.\footnote{Since the kernel is zonal (i.e.~symmetric about the north pole), its spherical harmonic coefficients are zero for any order $m\ne0$. For brevity of notation, we have therefore dropped the dependence on $m$ in $\kl$.} 
As can be seen in \cref{fig:kernel}(b), the coefficients $\kl$ decay rapidly and are approximately zero for $\ell \ge 7$, indicating that the kernel is effectively bandlimited.
\Cref{fig:kernel}(c) shows a 1D slice of the kernel computed using the first seven degrees of spherical harmonics overlaid on a 1D slice of the original kernel. The bandlimited approximation (red line) closely follows the original kernel (black line), indicating that the kernel is well approximated using this bandlimit. 

Technically speaking, the kernel is not strictly bandlimited, and hence the accuracy of the approximation depends on the chosen bandwidth. 
Naturally, a higher bandwidth yields a more accurate approximation of the kernel at the cost of requiring more spherical harmonic bases. This trade-off is illustrated in the first two columns of \cref{tab:bandwidth}. Here, the bandwidth $L$ denotes the number of spherical harmonic degrees used in the approximation.
In this work, we approximate the kernel using the bandwidth $L = 7$, which captures more than $99.9\%$ of the kernel's total energy.

Since the kernel is (approximately) bandlimited, the irradiance function also has the same bandwidth.
Therefore, we can approximate the irradiance function as a weighted sum of the first seven degrees of spherical harmonic bases:
\begin{equation}
    \En \approx \sum_{\ell = 0}^6 \sum_{m=-\ell}^{\ell} \Elm \, \Yn, \label{eq:E-sph-decomp-bandlimited}
\end{equation}
where $\Elm$ is the spherical harmonic coefficient of degree $\ell$ and order $m$, and $\Ylm$ is the corresponding real spherical harmonic basis function. 
In general, there are $L^2$ spherical harmonics in the first $L$ degrees. Therefore, the approximation in \cref{eq:E-sph-decomp-bandlimited} is a weighted sum of $49$ spherical harmonic bases. However, we know $\kl=0$ for odd degrees $\ell > 1$, and hence the convolution theorem~\cite{driscollComputingFourierTransforms1994} implies that the spherical harmonic coefficients $\Elm$ of those degrees are zero as well.
Therefore, \cref{eq:E-sph-decomp-bandlimited} reduces to a weighted sum of 31 spherical harmonics (third column of \cref{tab:bandwidth}). 
\textit{In other words, a good approximation of the irradiance function for any environment has just 31 degrees of freedom.}

\begin{table}[t]
    \caption{\textbf{Impact of the kernel's approximate bandwidth.} 
    While the first three degrees of spherical harmonics capture $99\%$ of the kernel's energy (second column)~\cite{basriLambertianReflectanceLinear2003,ramamoorthiSignalProcessingFrameworkInverse2001}, a higher bandwidth is needed to accurately approximate the irradiance function in real environments. 
    The number of degrees of freedom in the approximated irradiance function (third column) is the number of non-zero spherical harmonic coefficients in the first $L$ degrees.
    The error in the irradiance function (fourth column) is the error due to the bandwidth, as a percentage of the maximum irradiance in the scene, across 3,299 scenes.
    In this work, we approximate the kernel (and hence the irradiance function) using the first seven degrees of spherical harmonics (highlighted row).}
    \vspace{-0.1in}
    \centering
    \small
    \begin{tabular}{cccc}
        \toprule
        & Kernel & \multicolumn{2}{c}{Irradiance Function} \\
        \cmidrule(lr){2-2} \cmidrule(l){3-4}
        Bandwidth ($L$) & Energy Captured (\%) & D.O.F. & Approx. Error (\%) \\
        \midrule
        1 & $37.50$ & 1  & $25.23$ \\
        2 & $87.50$ & 4  & $9.08$  \\
        3 & $99.22$ & 9  & $1.77$  \\
        5 & $99.81$ & 18 & $0.77$  \\
        \chosenrow
        $\mathbf{7}$ & $\mathbf{99.92}$ & $\mathbf{31}$ & $\mathbf{0.44}$ \\
        9  & $99.96$ & 48 & $0.28$  \\
        11 & $99.98$ & 69 & $0.20$  \\
        \bottomrule
    \end{tabular}
    \vspace{-0.1in}
    \label{tab:bandwidth}
\end{table}

\vspace{-0.05in}
\subsection{Making the Measurements in the Spatial Domain}
One approach for sampling the irradiance function would be to directly measure the spherical harmonic coefficients $\Llm$ of the radiance function. Then, the coefficients $\Elm$ could be computed using the convolution theorem. 

\begin{figure}[t]
    \centering
    \includegraphics{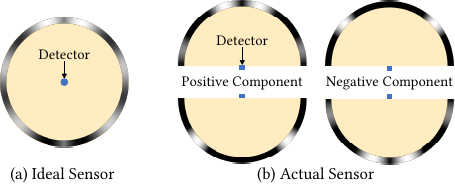}
    \vspace{-0.1in}
    \caption{
        \textbf{A spherical harmonic sensor?} (a)~One approach to sensing the irradiance function would be to directly measure its spherical harmonic coefficients. Each coefficient could be measured using a detector and mask whose transmittance is given by the corresponding spherical harmonic. (b)~In practice, this would require four detectors to measure a single spherical harmonic coefficient, since each detector can only receive light from  a hemisphere, and a mask can only have positive transmittance values. This approach would require 124 detectors and hemispherical masks, while the  irradiance function estimation problem only has 31 degrees of freedom.
    }
    \label{fig:measuring-sph}
    \vspace{-0.15in}
\end{figure}

This approach, however, is impractical for several reasons. 
First, computing a single coefficient requires a detector that views the scene through a spherical amplitude mask, whose spatially varying transmittance function is given by the  corresponding spherical harmonic (see \cref{fig:measuring-sph}(a)).
Spherical masks, however, are difficult to fabricate. Second, the detector must be infinitesimally small such that the entire mask is viewed from essentially a single point. Third, a single detector only sees one hemisphere of the mask, and hence two detectors with hemispherical masks are needed to compute the result for the spherical mask. Furthermore, an amplitude mask can only have positive transmittance values, while spherical harmonics have both positive and negative values. Therefore, the measurement for each hemisphere would require the use of two hemispherical masks, one for the positive component and the other for the negative component. Hence, the measurement of each spherical harmonic coefficient would require the use of four masks and detectors, as shown in  \cref{fig:measuring-sph}(b). In short, measuring 31 coefficients would require 124 hemispherical masks and detectors.

Clearly, directly measuring the spherical harmonic coefficients is impractical. However, our analysis has revealed that a good approximation of the irradiance function only has 31 degrees of freedom. We therefore take a simpler approach of directly sampling the irradiance function in the spatial domain. This can be done by placing bare detectors (without masks) on the surface of a sphere. Given that so few detectors are needed, we can use large detectors that cover the entire surface of the sphere, as shown in \cref{fig:what-is}(d). The large detectors collect more light, allowing them to produce measurements with a high dynamic range and signal-to-noise ratio (SNR), and harvest energy to power the camera.

Although this analysis shows that 31 detectors (samples) are sufficient to reconstruct the irradiance function, we chose to use 49 detectors to ensure that the reconstruction from the measurements is an overdetermined system and hence less sensitive to noise.

\vspace{-0.05in}
\subsection{Where Should the Samples Be?}
\begin{figure}[t]
    \centering
    \includegraphics{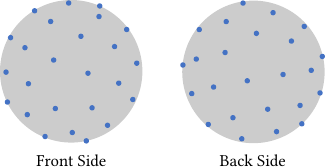}
    \vspace{-0.1in}
    \caption{
        \textbf{Sampling the irradiance function in the spatial domain.} Shown here is a roughly uniform arrangement of 49 samples (detectors) on the sphere, which is given by an approximate solution to the Tammes problem. 
    }
    \label{fig:sample-locations}
    \vspace{-0.1in}
\end{figure}

Now that we have decided to sample the irradiance function in the spatial domain, we must determine the sample locations. The problem of evenly arranging points on a sphere was first formalized by \citet{tammesOriginNumberArrangement1930}: Given $N$ points on the surface of a sphere, how should they be arranged to maximize the minimum distance between any two points? This problem, known as the Tammes problem, is well-studied~\cite{conwaySpherePackingsLattices1999}, and various sources provide tabulated solutions that are approximately optimal~\cite{sloaneTablesSphericalCodes1994, cohnTableSphericalCodes2024}. We use a solution for $N=49$ to determine the locations of our detectors, which are shown in \cref{fig:sample-locations}.

\vspace{-0.05in}

\subsection{Reconstructing the Irradiance Function}
Once we have sampled the irradiance function, we need to reconstruct it. To this end, the sampling process can be written as a linear system by evaluating the irradiance function in \cref{eq:E-sph-decomp-bandlimited} at the sample locations:
\begin{equation}
    \begin{bmatrix}
        E(\vn_1) \\
        \vdots \\
        E(\vn_{49})
    \end{bmatrix} = 
    \begin{bmatrix}
        Y_{0,0}(\vn_1) & \hdots & Y_{6,6}(\vn_1) \\
        \vdots & \ddots & \vdots \\
        Y_{0,0}(\vn_{49}) & \hdots & Y_{6,6}(\vn_{49})
    \end{bmatrix}
    \begin{bmatrix}
        E_{0,0} \\
        \vdots \\
        E_{6,6}
    \end{bmatrix}, \label{eq:sampling}
\end{equation}
where $\{\vn_1, \hdots, \vn_{49}\}$ is our set of 49 sample locations. 
\Cref{eq:sampling} forms an overdetermined system with 49 measurements and 31 unknowns (the non-zero spherical harmonic coefficients). We solve for the 31 coefficients using a constrained least squares solver to enforce the constraint that the irradiance function must be non-negative.
The supplemental material includes details of this reconstruction method.

\vspace{-0.05in}
\subsection{Simulations with Real Environment Maps}
\begin{figure}[t]
    \centering
    \includegraphics{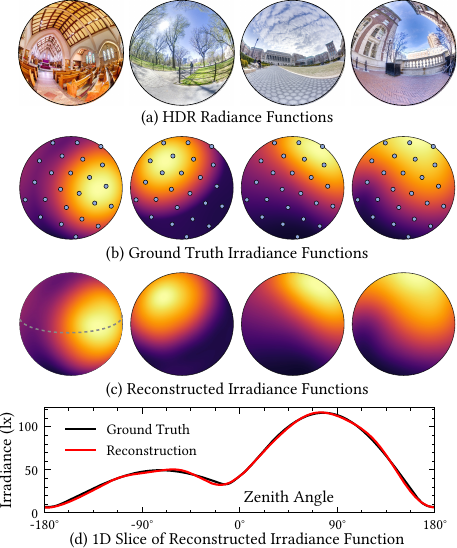}
    \vspace{-0.1in}
    \caption{
        \textbf{Simulations with real environment maps.}
        (a)~Example HDR radiance functions. Although the radiance function in each scene is a spherical image, we only show one hemisphere here.
        (b)~Corresponding ground truth irradiance functions, visualized using a colormap. We simulate the irradiance camera by sampling the irradiance function at the 49 sample locations (blue dots).
        (c)~The reconstructed irradiance functions computed from the 49 measurements. 
        (d)~A 1D slice of the reconstruction corresponding to the dashed gray line in~(c). The 49 measurements are sufficient for accurately reconstructing the irradiance function across diverse  real environments.
    }
    \vspace{-0.1in}
    \label{fig:simulations}
\end{figure}

We have simulated our approach for sampling and reconstructing the irradiance function using real environment maps from the Laval Indoor~\cite{gardnerLearningPredictIndoor2017} and UrbanSky~\cite{klotzMinimalSensingOrienting2025} datasets. These datasets together contain 3,299 high dynamic range images (radiance functions) taken in a variety of real indoor and outdoor scenes. \Cref{fig:simulations}(a) shows a few of the images from the two datasets, and \cref{fig:simulations}(b) shows the corresponding ground truth irradiance functions.

First, we evaluate the accuracy of the bandlimited approximation to the irradiance function. For each scene, we compute the bandlimited irradiance function using \cref{eq:E-sph-decomp-bandlimited} and compare it to the ground truth. The error in each scene is then computed as the root mean square error in the bandlimited irradiance function as a percentage of the maximum irradiance in that scene. The fourth column in \cref{tab:bandwidth} shows the average error across all 3,299 scenes due to the bandlimited approximation for different bandwidths $L$.  Our chosen bandwidth of $L=7$ yields an average error of $0.44\%$.

Next, we evaluate the accuracy of the irradiance function reconstructed from the 49 measurements.
To simulate an irradiance camera, we assume that each detector produces a measurement proportional to its irradiance. We sample the irradiance functions at the 49 locations (blue dots overlaid in \cref{fig:simulations}(b)). In addition, we emulate the noise characteristics and dynamic range of a real sensor by adding Gaussian noise and applying a saturation function. Finally, we use the samples to reconstruct the irradiance function and compute the error with respect to ground truth. \Cref{fig:simulations}(c) shows the reconstructed irradiance functions for each scene in \cref{fig:simulations}(a). \Cref{fig:simulations}(d) shows a 1D slice  through one of the reconstructions, where the slice corresponds to the gray dashed line in the first reconstructed irradiance function in \cref{fig:simulations}(c). On average, across all 3,299 scenes, the error between the reconstructed and ground truth irradiance functions was $1.41\%$ with respect to the maximum irradiance in each scene. These results show that our approach for sampling the irradiance function yields accurate reconstructions across diverse environments. Please see the supplemental material for simulation details and additional results.

\section{FluxCam: Irradiance Camera Prototype}

We have built FluxCam, a prototype irradiance camera, using 49 detectors arranged on the surface of a sphere.

\vspace{-0.05in}

\subsection{Camera Architecture}
\label{subsec:camera}
Each detector is a $25\times20\,\unit{\mm\squared}$ photovoltaic cell (Panasonic AM-5610), one of which is shown in \cref{fig:prototype}(a). The photovoltaic is connected to a custom-designed sensor board that switches the photovoltaic between a transimpedance amplifier (OPA607) for measuring irradiance and an energy harvester (BQ25570) to power the camera. 
All 49 sensor boards are connected to a custom-designed processor board (\cref{fig:prototype}(b-c)). The processor board includes a microcontroller (STM32L051C6) for communicating with the sensor boards, an RF transmitter (TI CC1310) to wirelessly transmit the measurements, and a supercapacitor to store energy harvested by the photovoltaics.

The camera is typically asleep, with all of the photovoltaics being used for energy harvesting. To measure irradiance, the sensor board switches its photovoltaic into the transimpedance amplifier, waits for the photocurrent to settle, and then digitizes the corresponding voltage using a 16-bit analog-to-digital converter (ADC). Once the measurement is made, the photovoltaic is immediately switched back into the energy harvester. To produce a single frame (49 irradiance measurements),  the processor board wakes up, sequentially reads out all 49 sensors as described above, wirelessly transmits the measurements, and then returns to sleep. The supplemental material includes further details of the camera architecture. 
In addition, the circuit schematics, PCB layout, and firmware are available online~\cite{irradiance-camera-web}.

\vspace{-0.05in}

\subsection{Rotating Detectors to Minimize Camera Size}
The detectors (photovoltaics) are placed on the surface of a sphere to sample the irradiance function at the 49 locations shown in \cref{fig:sample-locations}. Although the detector locations are fixed, each of them can be arbitrarily rotated about the axis that passes through its center and the center of the sphere without affecting its measured irradiance. This raises a practical question: how should the detectors be rotated such that they fit on the smallest possible sphere without overlapping?

\begin{figure*}[t]
    \centering
    \includegraphics[]{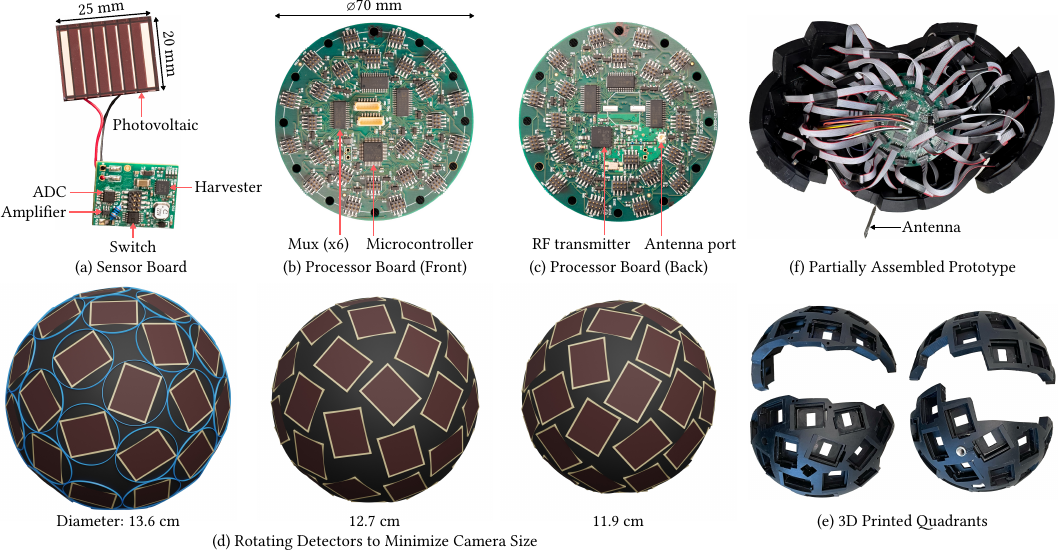}
    \vspace{-0.25in}
    \caption{
        \textbf{FluxCam: irradiance camera prototype.}
        (a)~Each sensor board consists of a photovoltaic attached to a custom circuit. The photovoltaic is used both for measuring irradiance and harvesting energy. 
        The (b)~front side and (c)~back side of the processor board show $49\times$ connectors for the sensor boards, readout circuitry, and an RF transmitter.
        (d)~We have optimized the rotations of the detectors to minimize the camera's size. Starting with a large sphere such that the bounding circles (blue) around each detector (dark brown) and its necessary padding (gold) do not overlap, we iteratively adjust the detector rotations and shrink the sphere until the detectors cannot be rotated without overlap. This yields a sphere with a smaller diameter.
        (e)~We 3D printed the final design in four quadrants using stereolithography.
        (f)~The inside of the camera. {\bf The supplemental video includes a time-lapse clip of the camera assembly.}
    }
    \label{fig:prototype}
    \vspace{-0.1in}
\end{figure*}

As a baseline, consider the smallest sphere that is large enough to avoid detector overlap, regardless of the rotations of the detectors. 
This case is illustrated on the left side of \cref{fig:prototype}(d), in which the detector rotations are chosen at random, and the sphere is large enough such that the bounding circles (blue) do not overlap. A $1\,\unit{\mm}$ padding (gold border) is added to the dimensions ($25\times20\,\unit{\mm\squared}$) of each detector to ensure a minimum separation between detectors in the final design. This conservative approach to packing the detectors on the sphere yields a sphere with a diameter of $13.6\,\unit{\cm}$.

To obtain a more compact design, we developed a greedy algorithm to optimize the rotations of the detectors. Beginning with the conservatively sized sphere described above, we adjust the rotation of each detector to maximize its minimum distance from its neighboring detectors. Then, we decrement the diameter of the sphere by a small amount ($1\,\unit{\mm}$) and repeat this process. This approach to shrinking the sphere is repeated until the sphere is too small to pack all of the detectors (inclusive of their paddings) without overlap. This method reduced the diameter of the sphere from $13.6\,\unit{\cm}$ to $11.9\,\unit{\cm}$ (\cref{fig:prototype}(d)). To add space for additional surface artifacts (e.g.,~mounting holes and debugging wires), we slightly increased the sphere's diameter to $12.8\,\unit{\cm}$. We then 3D printed this design as four quadrants using stereolithography (see \cref{fig:prototype}(e)).

\Cref{fig:prototype}(f) shows the inside of the camera during the assembly process. {\bf The supplemental video shows how FluxCam was assembled.} The processor board is held at the center of the camera, and each photovoltaic fits into a slot on the outside of the shell. Each sensor board is mounted inside the shell (behind the photovoltaics) and has a cable that connects it to the processor board. The antenna is placed outside the camera to maximize the communication range. \Cref{fig:teaser}(a) shows the fully assembled camera. 

\vspace{-0.05in}

\subsection{Radiometric Performance}
\begin{figure*}[t]
    \centering
    \includegraphics[]{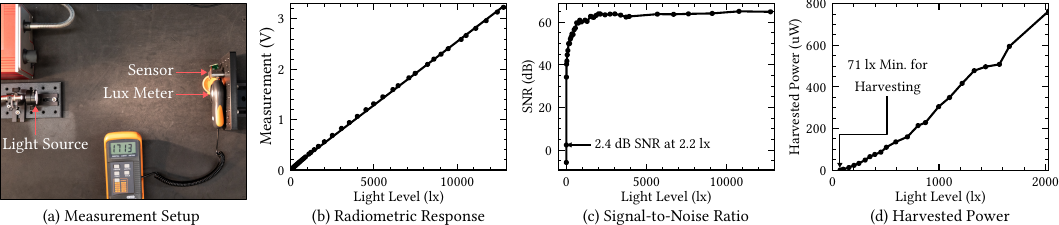}
    \vspace{-0.25in}
    \caption{
        \textbf{Sensor performance.}
        (a)~Setup used to measure the radiometric characteristics of the sensor.
        (b)~The radiometric response is linear over the entire dynamic range (up to $13{,}000\,\unit{\lux}$).
        (c)~SNR of the sensor. At just $2.2\,\unit{\lux}$, the SNR is $2.4\,\unit{\dB}$, which is near the minimum detectable signal. This corresponds to a wide dynamic range of $75.4\,\unit{\dB}$.
        (d)~Power harvested by the sensor. Due to the harvester's overhead, the sensor board requires at least $71\,\unit{\lux}$ to harvest power.
    }
    \label{fig:performance}
\end{figure*}

We have characterized the performance of our sensor using the measurement setup in \cref{fig:performance}(a). A controlled halogen light source (Thorlabs OSL2) was used to illuminate the detector head-on, and a lux meter placed next to the detector was used to measure the ground truth illuminance.

The measurements made by FluxCam are weighted by the spectral response of the photovoltaics. In our case, the photovoltaics have a spectral response that closely resembles the photopic response of the human eye~\cite{photovoltaic-datasheet}. Due to this spectral response, FluxCam measures illuminance (in lux) and hence can be calibrated using a lux meter. If the spectral distribution of the incident illumination is known, then irradiance can be computed from the measured illuminance using a scale factor that can be determined via calibration.

\Cref{fig:performance}(b) shows the sensor's radiometric response function. The sensor is capable of measuring light levels as high as $13{,}000\,\unit{\lux}$, and the response function is linear over the entire dynamic range. \Cref{fig:performance}(c) shows the SNR of the sensor, which is higher than $60\,\unit{\dB}$ for light levels brighter than $750\,\unit{\lux}$. At just $2.2\,\unit{\lux}$, the SNR is $2.4\,\unit{\dB}$. If we use this light level as the minimum detectable signal, this corresponds to a remarkably wide dynamic range of $75.4\,\unit{\dB}$. 
This performance can be attributed to the detector's large active area and the absence of any attenuating optics,
allowing the detector to collect substantially more light than a conventional pixel.
Details of the sensor performance measurements are provided in the supplemental material.

\begin{figure*}[t]
    \centering
    \includegraphics[]{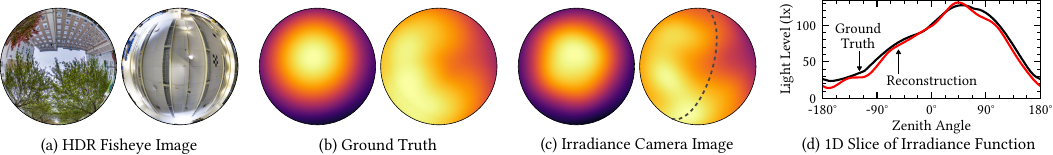}
    \vspace{-0.1in}
    \caption{
        \textbf{Reconstructed irradiance functions using FluxCam.}
        (a)~HDR images of two real scenes. (b)~The corresponding ground truth irradiance functions.
        (c)~Reconstructed irradiance functions using the 49 measurements produced by FluxCam in each scene. Across four different scenes, the measured irradiance functions attain an average RMSE of $3.93\%$ with respect to the maximum irradiance in each scene.
        (d)~Reconstruction accuracy for the 1D slice corresponding to the dashed gray line in~(c).
    }
    \label{fig:real-irradiance}
    \vspace{-0.1in}
\end{figure*}

\vspace{-0.05in}

\subsection{Self-Powered Operation}
As described in Section \ref{subsec:camera}, in addition to measuring irradiance, the 49 sensors also harvest energy to power FluxCam. \Cref{fig:performance}(d) shows the power harvested by a single sensor as a function of light level. The plot does not pass through the origin since the energy harvester circuit itself consumes a small amount of power. As a result, the sensor requires a minimum illumination of $71\,\unit{\lux}$ to harvest energy. This threshold is a limitation of the harvester circuit and is independent of the sensor's dynamic range. As described before, all sensors can measure light levels as low as $2.2\,\unit{\lux}$. In environments for which the average illumination falling on the camera is roughly $1{,}000\,\unit{\lux}$, the total energy harvested by all 49 sensors is sufficient to operate the camera at 30 FPS. In the supplemental material, we have included details of the experimental setup and an algorithm to adaptively adjust the camera's framerate based on the harvested power (i.e.,~the brightness of the environment).

\vspace{-0.05in}

\subsection{Accuracy of Reconstructed Irradiance Functions}
We have evaluated the accuracy of the irradiance functions reconstructed using FluxCam by comparing them with ones computed using HDR spherical images (taken with a Ricoh Theta Z1) in four indoor and outdoor scenes. Two of these HDR images are shown in 
\cref{fig:real-irradiance}(a), and their corresponding irradiance functions (ground truth) are shown in \cref{fig:real-irradiance}(b).
The irradiance functions reconstructed from the 49 measurements made by FluxCam  are shown in \cref{fig:real-irradiance}(c). The reconstruction accuracy of a 1D slice through one of the reconstructed functions can be seen in \cref{fig:real-irradiance}(d). Across the four scenes we measured, the average RMSE in the reconstructed irradiance function, as a fraction of the maximum irradiance in each scene, was $3.93\%$. The supplemental material includes results for all four scenes.

\vspace{-0.05in}

\section{Example Applications}

\subsection{Estimating Egomotion}
In our first application, we show that FluxCam can be used to estimate egomotion. In the case of a pure rotation of the camera, the irradiance function rotates by the same amount, and hence estimating the camera's rotation is straightforward. Real motion, however, includes translation, causing the irradiance function to undergo a complex, scene-dependent transformation. Therefore, we trained a network to predict the relative rotation between two camera poses, which includes an arbitrary translation between them, using the irradiance measurements at each pose. Training data was generated in simulation by rendering the irradiance of the camera's 49 detectors in a synthetic urban environment. For each training sample, we rendered the measurements at an initial camera pose, applied a random rotation and a small translation, and rendered the measurements again at the new pose. The camera was positioned at different distances from the ground to include the effects of parallax due to translation. Additionally, the rendering process accounts for the fact that each detector views the scene from a different viewpoint. Using this procedure, we generated two million pairs of irradiance measurements for training. We tested the network using $319{,}712$ pairs of simulated measurements (with rotation and translation), and found the network to yield an average rotation error of $2.25\unit{\degree}$. 

In practice, a camera's trajectory varies smoothly with time. Therefore, we can refine the estimated camera orientation using temporal filtering. \Cref{fig:experiments}(a) shows the estimated camera orientation for two scenarios: rolling FluxCam on the ground (left) and tossing FluxCam in the air (right). In both cases, the network first predicts the relative orientation between frames, after which temporal filtering is used to determine the camera's (relative) orientation. In effect, FluxCam functions as an untethered optical gyroscope. \textbf{Please see the supplemental video.} Details regarding the network architecture and the simulation results are provided in the supplemental material.

\vspace{-0.05in}

\subsection{Relighting using an Untethered Light Probe}
Mirror probes are widely used to measure the environmental illumination~\cite{debevecRenderingSyntheticObjects1998,reinhardHighDynamicRange2010,debevecSingleshotLightProbe2012}. While this approach can be used for relighting objects with complex appearances, the mirror must be fully visible to the camera. In other words, it cannot measure illumination in locations that are fully or partially occluded by other objects. Since FluxCam is untethered, it does not need line of sight with the camera. For example, on the left in \cref{fig:experiments}(b), FluxCam is partially occluded by the vase but measures and transmits the complete irradiance function at that location. Although the irradiance function has less detail than the radiance function, it is useful for diffuse relighting. On the right in \cref{fig:experiments}(b), the measured irradiance is used to render and insert a diffuse object (a light wood vase with grooves)
at the location where FluxCam was positioned.

Since FluxCam's measurements are weighted by the spectral response of the photovoltaics, the measurements do not reveal the illuminant's color. Therefore, in this example, we assume a white illuminant and perform relighting in the RGB color space. In the future, FluxCam could be modified to measure spectral irradiance using a larger number of detectors with spectral filters, enabling physically accurate relighting under different illuminants.

\vspace{-0.05in}

\subsection{Workspace Monitoring}

Many vision tasks only involve high-level inferences about the statistics of objects in a scene. These tasks, referred to as lightweight~\cite{klotzMinimalistVisionFreeform2025}, do not require the detection of fine-grained visual details. Although an irradiance camera produces few measurements, they have high precision and dynamic range and hence encode visual cues about the environment that can be used to solve lightweight tasks. As an example, consider the workspace in \cref{fig:experiments}(c). The goal is to determine which desks are occupied, detect when the door is open, and determine which lamps are on (dashed black boxes in the first image of \cref{fig:experiments}(c)). We placed FluxCam in the center of the room to solve these tasks. For each task, we trained classifiers using 23 minutes of labeled data. The classifier outputs for the test data are overlaid on the images in \cref{fig:experiments}(c), and the inset images show the irradiance functions measured by the camera. The system detects when the door is open with 90.9\% accuracy, the occupancy of all three desks with 87.6\% accuracy, and the activation of the lamps with 99.9\% accuracy. \textbf{Please see the supplemental video.} In addition to being self-powered and untethered, the irradiance camera preserves privacy in such a setting since it does not capture the visual details needed to identify individuals.

\vspace{-0.05in}

\subsection{Omnidirectional Pyranometer}
\begin{figure*}[p]
    \centering
    \includegraphics{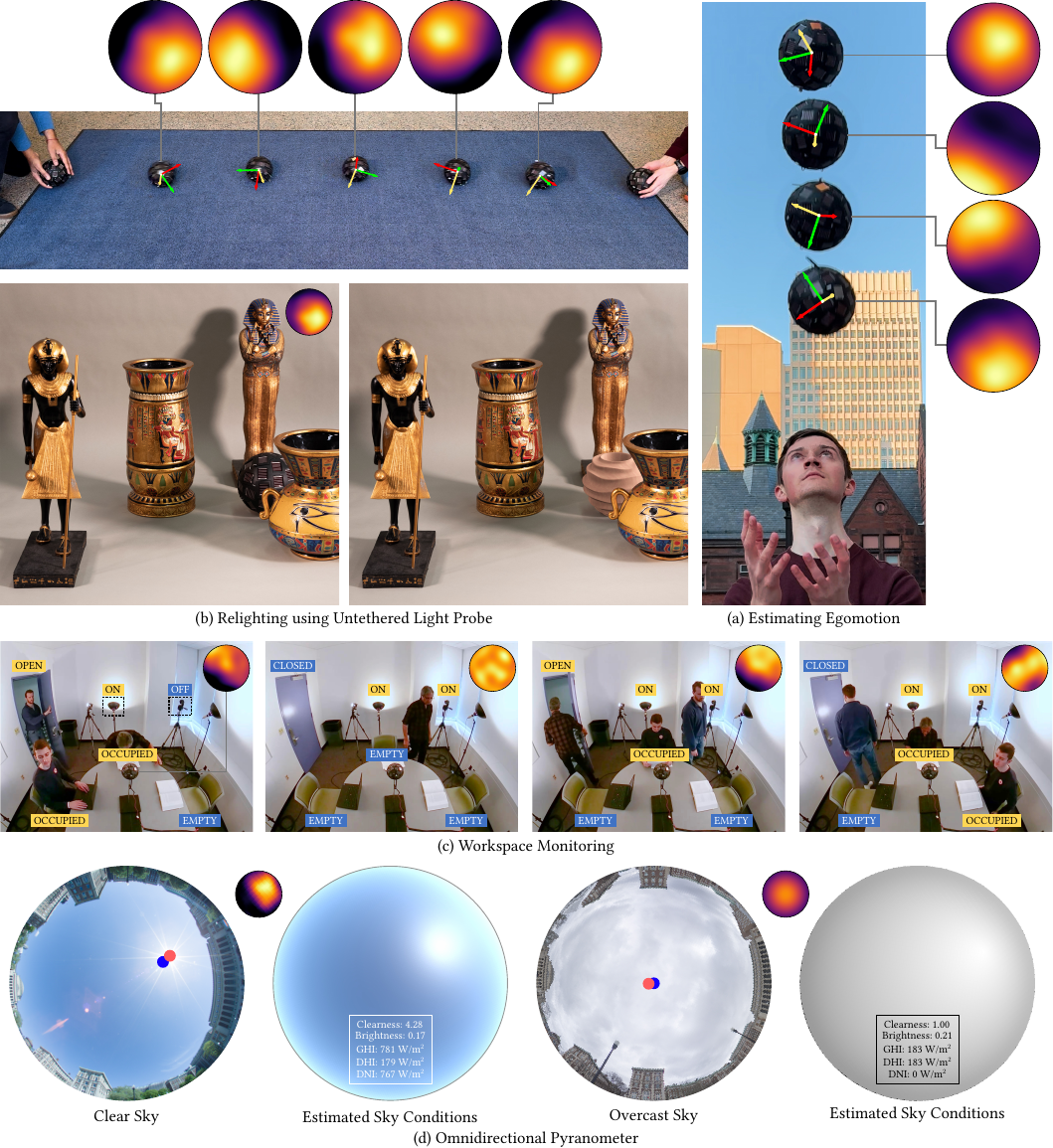}
    \vspace{-0.25in}
    \caption{
        \textbf{Example applications of an irradiance camera.}
        (a)~FluxCam can be used as an optical gyroscope to determine the camera's orientation. Here we show the computed orientation as the camera is rolled across a floor and tossed in the air.
        (b)~The camera can serve as an untethered light probe for diffuse relighting. We placed the camera in a location that is partially occluded by another object (left image) and measured the irradiance function (inset image). This allows us to relight and insert a diffuse object (right image).
        (c)~FluxCam can be used to monitor a workspace. Here we use the camera's 49 measurements to determine which desks are occupied, detect when the door is open, and determine which lights are on (dashed black boxes in the left image).
        (d)~FluxCam can be used as an omnidirectional pyranometer to estimate the best orientation of a solar panel (red dots). Additionally, the measurements can be used to reconstruct the sky, allowing us to quantify the current sky conditions. \textbf{Please see the supplemental video.}
    }
    \label{fig:experiments}
\end{figure*}

Pyranometers are widely used in meteorology and solar energy to measure the irradiance of a surface with a fixed orientation~\cite{duffieSolarEngineeringThermal2006}. The irradiance camera is, in effect, an omnidirectional pyranometer. Consider the HDR fisheye image on the left of \cref{fig:experiments}(d), which was taken on a clear day. We placed FluxCam in this environment to measure the irradiance function (small image outside the fisheye image). These measurements can be used to compute the direction of maximum irradiance (red dot), which corresponds to the orientation of a solar panel that would maximize the harvested energy. For comparison, the direction computed from the fisheye image is shown as a blue dot.

An irradiance camera can also be used to determine the sky conditions. To this end, we fit our irradiance measurements to a low-dimensional analytical sky model~\cite{perezAllweatherModelSky1993}, which allows us to estimate the sky's brightness and clearness. Furthermore, we can use the fitted sky model to decompose the irradiance into components due to the sun and the sky, which is typically parameterized by the global horizontal irradiance (GHI), diffuse horizontal irradiance (DHI), and direct normal irradiance (DNI)~\cite{duffieSolarEngineeringThermal2006}. \Cref{fig:experiments}(d) shows reconstructed skies under clear and overcast conditions using the Hosek--Wilkie and Perez models, respectively~\cite{hosekAnalyticModelFull2012,perezAllweatherModelSky1993}. 
In this example, we do not use explicit spectral measurements to fit the analytical sky model. Since the spectral response of FluxCam's detectors closely resembles the photopic response, it measures illuminance (in $\unit{\lux}$), which we convert to irradiance (in $\unit{\watt\per\square\meter}$) using a luminous efficacy of $109\,\unit{\lumen\per\watt}$ before fitting. 
For the overcast case, the irradiance measurements were simulated using an HDR spherical image from the UrbanSky dataset. The supplemental material includes a quantitative evaluation of this approach to sky condition estimation.

\vspace{-0.1in}
\section{Discussion}
We have introduced the concept of an irradiance camera and demonstrated its potential applications. There are several directions we plan to pursue as future work. First, 
our current prototype operates at 30 FPS when the average illumination received by it is roughly $1{,}000\,\unit{\lux}$. The camera's power consumption can be further optimized so that it can operate in dimly lit indoor environments.
Second, the camera can be miniaturized by using smaller photovoltaics. This would reduce the harvested energy and hence the framerate of the camera, but the lower framerates would be adequate for a variety of applications. Third, while we have shown that the camera can be used to estimate rotation, we believe our approach can be used to estimate the direction of translation as well, making it useful for self-powered egomotion estimation on a variety of robotic platforms, such as drones and wheeled robots. Finally, while our current implementation ensures each detector has a hemispherical field of view, we believe that using apertures to reduce the field of view would produce measurements that are richer in information for certain lightweight tasks. With these improvements, we hope that the irradiance camera could serve as a powerful untethered sensor for a wide variety of vision applications.

\begin{acks}
This work was supported by the Office of Naval Research (ONR) under award number N00014-23-1-2096 and the National Science Foundation (NSF) and Center for Smart Streetscapes (CS3) under NSF Cooperative Agreement No. EEC-2133516. Jeremy Klotz was supported by a National Defense Science and Engineering Graduate (NDSEG) Fellowship. The authors are grateful to Behzad Kamgar-Parsi for his support and encouragement. The authors also thank Mikhail Fridberg for help designing the camera electronics, Silvia Sellán for discussions about Blender, Joaquin Palacios for help with 3D printing, Junbang Liang for discussions about transformers, and Matthew Beveridge for technical feedback.

\end{acks}

\bibliographystyle{ACM-Reference-Format}
\bibliography{irradiance-camera}

\clearpage
\setcounter{section}{0}
\renewcommand{\thesection}{S\arabic{section}}
\renewcommand{\thesubsection}{S\arabic{section}.\arabic{subsection}}
\renewcommand{\thesubsubsection}{S\arabic{section}.\arabic{subsection}.\arabic{subsubsection}}
\setcounter{figure}{0}
\renewcommand{\thefigure}{S\arabic{figure}}
\setcounter{table}{0}
\renewcommand{\thetable}{S\arabic{table}}
\renewcommand{\theHsection}{S\arabic{section}}
\renewcommand{\theHfigure}{S\arabic{figure}}
\renewcommand{\theHtable}{S\arabic{table}}

\raggedbottom
\allowdisplaybreaks

\section{Spherical Harmonic Conventions}
Throughout the paper, we use the real spherical harmonics $\Ylm$, indexed by degree $\ell \ge 0$ and order $-\ell \le m \le \ell$. For any direction $\vn$ with zenith angle $\theta$ and azimuthal angle $\phi$, the real spherical harmonics are defined as
\begin{equation}
    \Ylm(\theta, \phi) =
    \begin{cases}
        \sqrt{2} \, N_\ell^{|m|} \, P_\ell^{|m|}(\cos \theta) \, \sin(|m| \phi), & m < 0, \\[6pt]
        N_\ell^{0} \, P_\ell^{0}(\cos \theta),                       & m = 0, \\[6pt]
        \sqrt{2} \, N_\ell^{m} \, P_\ell^{m}(\cos \theta) \, \cos(m \phi),       & m > 0,
    \end{cases}
    \label{eq:sph}
\end{equation}
where $P_\ell^m$ is the associated Legendre polynomial of degree $\ell$ and order $m \ge 0$, and $N_\ell^m$ is the normalization constant
\begin{equation}
    N_\ell^m = (-1)^m \sqrt{\frac{2\ell + 1}{4\pi} \frac{(\ell - m)!}{(\ell + m)!}}.
    \label{eq:sph-norm}
\end{equation}
We use the Condon--Shortley phase convention in which the $(-1)^m$ phase term is included in $P_\ell^m$, which cancels out with the $(-1)^m$ phase term in $N_\ell^m$.

\section{Reconstructing the Irradiance Function}

\subsection{Reconstruction Method}
We reconstruct the spherical harmonic coefficients of the irradiance function by inverting the sampling process in eq.~(4) in the paper. Recall from eq.~(3) in the paper that only $31$ of the $49$ coefficients in the first seven degrees are non-zero. We therefore solve only for those $31$ coefficients, whose degrees and orders we index as $(\ell_1, m_1), \hdots, (\ell_{31}, m_{31})$. Since irradiance must be non-negative, not every set of spherical harmonic coefficients corresponds to a physically valid irradiance function. To enforce this constraint, we solve for the 31 coefficients from the 49 measurements using constrained least squares:
\begin{align}
\min_{x} \quad &
\left\|
    Ax - b
\right\|_2^2 \\
\textrm{subject to} \quad & \nonumber
Gx \ge 0,
\end{align}
where
\begin{align}
A &=
\underbrace{
\begin{bmatrix}
        Y_{\ell_1 m_1}(\vn_1) & \hdots & Y_{\ell_{31} m_{31}}(\vn_1) \\
        \vdots & \ddots & \vdots \\
        Y_{\ell_1 m_1}(\vn_{49}) & \hdots & Y_{\ell_{31} m_{31}}(\vn_{49})
\end{bmatrix}
}_{49 \times 31},\label{eq:A} \\
x &=
\underbrace{
\begin{bmatrix}
        E_{\ell_1 m_1} \\
        \vdots \\
        E_{\ell_{31} m_{31}}
\end{bmatrix}
}_{31 \times 1}, \\
b &=
\underbrace{
\begin{bmatrix}
        E(\vn_1) \\
        \vdots \\
        E(\vn_{49})
\end{bmatrix}
}_{49 \times 1}, \\
G &=
\underbrace{
\begin{bmatrix}
        Y_{\ell_1 m_1}(\vn'_1) & \hdots & Y_{\ell_{31} m_{31}}(\vn'_1) \\
        \vdots & \ddots & \vdots \\
        Y_{\ell_1 m_1}(\vn'_K) & \hdots & Y_{\ell_{31} m_{31}}(\vn'_K)
\end{bmatrix}
}_{K \times 31}.
\end{align}
Here, $\{\vn'_1, \hdots, \vn'_K\}$ is a dense set of uniformly distributed directions on the sphere generated using HEALPix~\cite{gorskiHEALPixFrameworkHighResolution2005}. This constrained least squares formulation ensures that the reconstructed irradiance function evaluated at other points on the sphere, not just the measured points $\{\vn_1, \hdots, \vn_{49}\}$, is non-negative. In our implementation, we use $K=4800$ points. We solve for the spherical harmonic coefficients using the \texttt{quadprog} solver in \texttt{qpsolvers}~\cite{qpsolvers}.

The 49 sample locations $\{\vn_1, \hdots, \vn_{49}\}$ are listed in \cref{tab:sample-coords}. Code for reconstructing the irradiance function from the 49 samples is available online~\cite{irradiance-camera-web}.

\begin{table}[t]
    \caption{\textbf{Sample locations.} The coordinates of the 49 sample locations $\{\vn_1, \hdots, \vn_{49}\}$ on the unit sphere. These samples are an approximate solution to the Tammes problem~\cite{sloaneTablesSphericalCodes1994}.}
    \vspace{-0.05in}
    \label{tab:sample-coords}
    \centering
    \setlength{\tabcolsep}{4pt}
    \small
    \begin{tabular}{crrr >{\hspace{1.5em}}c<{\hspace{1.5em}} crrr}
        \toprule
        \multicolumn{1}{c}{$n$} & \multicolumn{1}{c}{$X$} & \multicolumn{1}{c}{$Y$} & \multicolumn{1}{c}{$Z$} & &
        \multicolumn{1}{c}{$n$} & \multicolumn{1}{c}{$X$} & \multicolumn{1}{c}{$Y$} & \multicolumn{1}{c}{$Z$} \\
        \midrule
        1  & $0.407$  & $0.511$  & $-0.757$ & & 26 & $0.956$  & $-0.247$ & $0.157$  \\
        2  & $-0.569$ & $-0.802$ & $0.179$  & & 27 & $-0.131$ & $0.967$  & $0.217$  \\
        3  & $0.307$  & $0.865$  & $-0.396$ & & 28 & $-0.840$ & $0.042$  & $0.540$  \\
        4  & $0.654$  & $-0.289$ & $0.699$  & & 29 & $-0.263$ & $-0.761$ & $0.593$  \\
        5  & $-0.710$ & $-0.288$ & $-0.643$ & & 30 & $0.403$  & $-0.915$ & $-0.012$ \\
        6  & $-0.876$ & $-0.398$ & $0.272$  & & 31 & $0.046$  & $0.213$  & $-0.976$ \\
        7  & $-0.484$ & $0.276$  & $0.831$  & & 32 & $-0.096$ & $-0.991$ & $0.097$  \\
        8  & $0.512$  & $0.015$  & $-0.859$ & & 33 & $-0.194$ & $0.936$  & $-0.295$ \\
        9  & $0.823$  & $-0.223$ & $-0.523$ & & 34 & $-0.591$ & $0.656$  & $-0.470$ \\
        10 & $-0.396$ & $-0.866$ & $-0.304$ & & 35 & $-0.583$ & $0.812$  & $0.022$  \\
        11 & $0.726$  & $0.668$  & $-0.166$ & & 36 & $0.373$  & $0.921$  & $0.113$  \\
        12 & $-0.305$ & $-0.163$ & $-0.938$ & & 37 & $0.507$  & $0.391$  & $0.768$  \\
        13 & $-0.801$ & $-0.559$ & $-0.213$ & & 38 & $-0.909$ & $0.416$  & $-0.037$ \\
        14 & $0.013$  & $0.401$  & $0.916$  & & 39 & $0.780$  & $-0.601$ & $-0.174$ \\
        15 & $-0.289$ & $0.714$  & $0.638$  & & 40 & $-0.080$ & $0.680$  & $-0.729$ \\
        16 & $0.286$  & $-0.036$ & $0.958$  & & 41 & $0.221$  & $0.780$  & $0.586$  \\
        17 & $-0.597$ & $-0.382$ & $0.705$  & & 42 & $0.125$  & $-0.514$ & $0.849$  \\
        18 & $0.169$  & $-0.367$ & $-0.915$ & & 43 & $0.972$  & $0.235$  & $-0.027$ \\
        19 & $-0.438$ & $0.325$  & $-0.838$ & & 44 & $0.104$  & $-0.896$ & $-0.432$ \\
        20 & $0.807$  & $0.293$  & $-0.513$ & & 45 & $0.230$  & $-0.851$ & $0.472$  \\
        21 & $0.861$  & $0.136$  & $0.490$  & & 46 & $-0.988$ & $-0.086$ & $-0.125$ \\
        22 & $-0.727$ & $0.534$  & $0.432$  & & 47 & $0.500$  & $-0.617$ & $-0.608$ \\
        23 & $0.684$  & $-0.650$ & $0.331$  & & 48 & $0.713$  & $0.610$  & $0.347$  \\
        24 & $-0.828$ & $0.200$  & $-0.523$ & & 49 & $-0.236$ & $-0.634$ & $-0.737$ \\
        25 & $-0.218$ & $-0.146$ & $0.965$  & &    &          &          &          \\
        \bottomrule
    \end{tabular}
    \vspace{-0.05in}
\end{table}

\subsection{Reconstruction Stability}
\begin{figure}[t]
    \centering
    \includegraphics{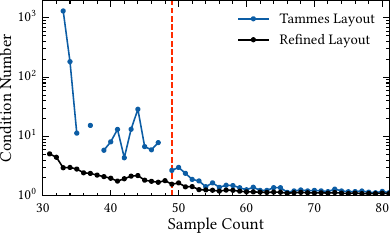}
    \vspace{-0.1in}
    \caption{
        \textbf{Conditioning of the reconstruction problem for different sample arrangements.} The condition number of $A$ using sample arrangements given by approximate solutions to the Tammes problem is shown in blue. The condition number of $A$ after iteratively refining the sample locations is shown in black.
    }
    \label{fig:cond-number}
    \vspace{-0.05in}
\end{figure}

We analyze the stability of the reconstruction method by examining the condition number of $A$ (\cref{eq:A}). The blue line in \Cref{fig:cond-number} plots the condition number of $A$ corresponding to roughly uniform sample arrangements (approximate solutions to the Tammes problem) for different sample counts. For the 49 samples used by our system (vertical red line), $A$ has a condition number of $2.69$. As expected, the conditioning improves if more samples are used.

Even though there are 31 unknowns (i.e.,~$A$ has 31 columns), not every arrangement of $N \ge 31$ samples yields a well-conditioned reconstruction problem. The gaps in \cref{fig:cond-number} denote sample arrangements that are ill-conditioned. This occurs when the specific sample layout exhibits symmetries that produce linearly dependent columns in $A$. However, the conditioning for any given number of samples could be improved using a non-uniform sample layout. To illustrate this, we have iteratively adjusted the sample locations using gradient descent to minimize the condition number. The resulting sample layouts (black line in \cref{fig:cond-number}) would improve the conditioning compared to the uniform sample layouts from approximate solutions to the Tammes problem. For the layout using $N = 49$ samples, the condition number would decrease from $2.69$ to $1.57$ using this refined, non-uniform layout. In practice, such non-uniform sample layouts come at the cost of requiring a larger sphere since some samples (detectors) would be more tightly packed.

\section{Simulated Sensor Model}
\begin{figure*}[t]
    \centering
    \includegraphics{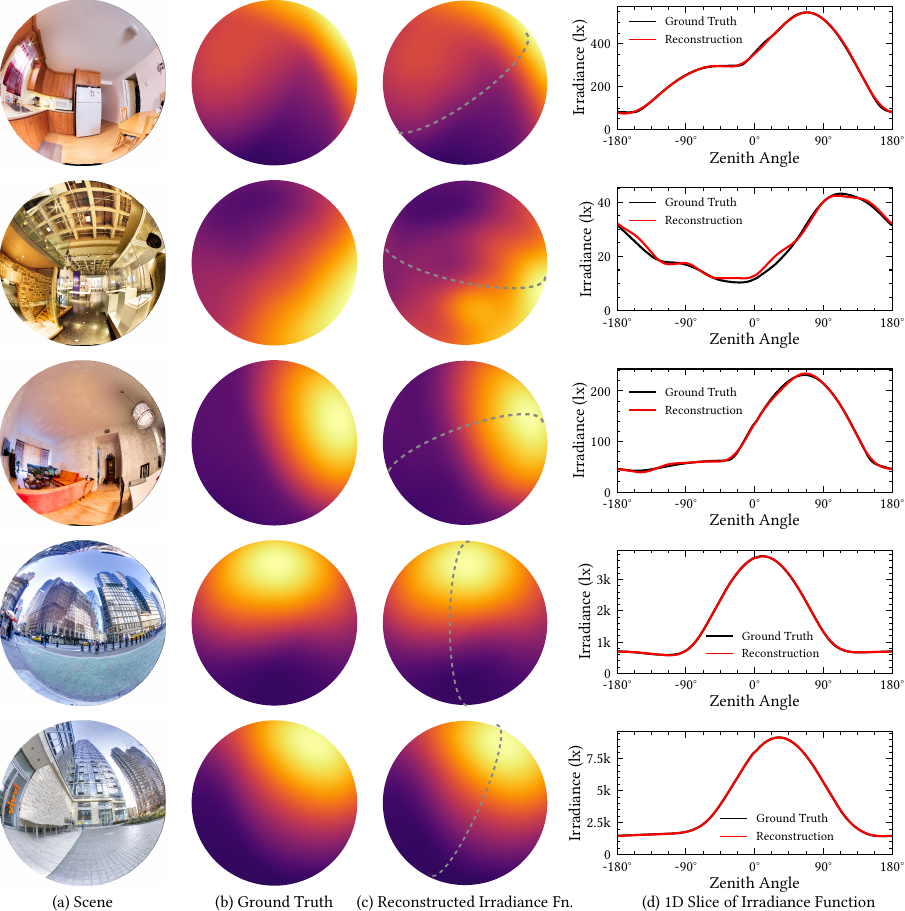}
    \caption{
        \textbf{Additional simulation results in scenes from the Laval Indoor and UrbanSky datasets.}
        (a)~HDR image of the scene (radiance function).
        (b)~Ground truth irradiance function.
        (c)~Reconstructed irradiance function from simulated measurements.
        (d)~1D slice of the reconstructed irradiance functions through the dashed gray lines in~(c).
    }
    \label{fig:sim-irradiance-supp}
\end{figure*}

For the experiments evaluating the accuracy of the reconstructed irradiance functions in simulation, we model a sensor that saturates at $20{,}000\,\unit{\lux}$ (corresponding to $3.3\,\unit{\volt}$) and includes $250\,\unit{\micro\volt}$ root mean square (RMS) Gaussian noise. For any scene in which the maximum illumination exceeds $20{,}000\,\unit{\lux}$ (i.e.,~the sensor would saturate), we simulate the effect of adding a 3-stop neutral density (ND) filter to all 49 detectors. Then, we model the saturation by clipping the measurements at $3.3\,\unit{\volt}$. \Cref{fig:sim-irradiance-supp} shows additional simulation results.

\section{FluxCam Architecture}

\Cref{tab:parts} lists the components used in FluxCam, our prototype irradiance camera. Each photovoltaic is connected through a switch to a transimpedance amplifier (TIA) with a gain of $4700\,\unit{\V / \A}$. The TIA measures the photovoltaic's short circuit current without applying any reverse bias. A 16-bit ADC measures the output of the TIA and sends that measurement to the processor board over an SPI bus. Once the measurement is made, the photovoltaic is switched back into the energy harvester to power the camera. \textbf{The circuit schematics, PCB layouts, and firmware are available online~\cite{irradiance-camera-web}.}

\begin{table}[t]
    \caption{\textbf{Components in the prototype irradiance camera.} The hardware design (circuit schematics and PCB layouts) is available online~\cite{irradiance-camera-web}.}
    \vspace{-0.05in}
    \centering
    \small
    \begin{tabular}{lcl}
        \toprule
        Component & Quantity & Description \\
        \midrule
        Photovoltaic & 49 & Panasonic AM-5610CAR-DGK-T \\
        Amplifier & 49 & TI OPA607 \\
        ADC & 49 & Analog Devices AD7683 \\
        Energy Harvester & 49 & TI BQ25570 \\
        Switch & 49 & Analog Devices ADG787 \\
        \midrule
        Readout Microcontroller & 1 & STM32L051C6T6 \\
        RF Microcontroller & 1 & TI CC1310F128 \\
        Mux & 6 & Analog Devices ADG706 \\
        Supercapacitor & 1 & $107\,\unit{\milli\farad}$ \\
        Antenna & 1 & ANTX150P116B08683 \\
        \bottomrule
    \end{tabular}
    \label{tab:parts}
    \vspace{-0.05in}
\end{table}

\begin{figure}[t]
    \centering
    \includegraphics{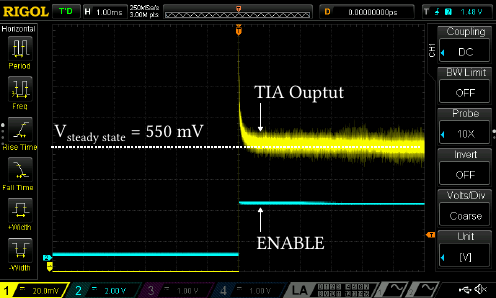}
    \vspace{-0.05in}
    \caption{
        \textbf{Settling time of transimpedance amplifier output.}
        When the ENABLE signal goes high, the sensor board switches the photovoltaic from the energy harvester into the transimpedance amplifier. The output of the transimpedance amplifier, which is proportional to the photovoltaic's short circuit current, takes roughly $20\,\unit{\milli\second}$ to settle to its steady state value ($550\,\unit{\milli\volt}$ in this case). The choice of when to sample the TIA output represents a tradeoff between producing high fidelity measurements and harvesting more energy.
    }
    \label{fig:scope-supp}
    \vspace{-0.1in}
\end{figure}

When the photovoltaic is initially switched into the TIA, the photocurrent requires time to settle before reaching its steady state current. \Cref{fig:scope-supp} shows the transient response of the TIA output immediately after switching the detector from the energy harvester into the TIA. The switch occurs at the moment the ENABLE signal goes high (see \cref{fig:scope-supp}). The transient takes roughly $20\,\unit{\milli\second}$ to achieve steady state. This settling time can be set in firmware to trade off measurement fidelity (which benefits from a long settling time) with energy harvesting efficiency (which benefits from a short settling time). 

To minimize the power consumption, the microcontrollers on the processor board are typically asleep. In our current implementation, they wake up at a predefined interval (nominally 30 FPS) to read out the measurements from the sensor boards, wirelessly transmit them, and then go back to sleep. At 30 FPS with a $50\,\unit{\micro\second}$ settling time, the processor board consumes $500\,\unit{\micro\ampere}$, and each sensor board consumes roughly $3\,\unit{\micro\ampere}$.

\section{FluxCam Performance}
\subsection{Radiometric Performance}
\begin{figure}[t]
    \centering
    \includegraphics{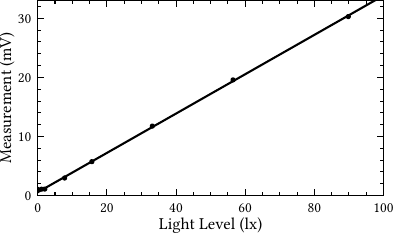}
    \vspace{-0.1in}
    \caption{\textbf{Radiometric response function at low light levels.}}
    \label{fig:response-supp}
    \vspace{-0.05in}
\end{figure}

To characterize the radiometric performance and signal-to-noise ratio (SNR), we used a long settling time of $20\,\unit{\milli\second}$. In this case, we operated the sensor at 30 FPS and attached a $681\,\unit{\kilo\ohm}$ load to the output of the harvester. To measure the radiometric response at a single light level, we used the average measurement taken over 15 seconds (corresponding to 450 measurements at 30 FPS). \Cref{fig:response-supp} shows the radiometric response function at very low light levels, with a line fit to the measurements. 

To measure the radiometric response and SNR at low light levels (roughly $1{,}500\,\unit{\lux}$ and below), we used a Thorlabs OSL2 halogen lamp. To measure the performance at higher light levels, we used a brighter halogen lamp (IBM 663533 B). We applied a constant scale factor to measurements taken under the brighter halogen lamp to account for the spectral mismatch with the dimmer lamp.

Each of the 49 detectors may respond differently to the same light level due to component tolerances. To account for this, we measured the radiometric response of all 49 detectors at two different light levels using a controlled light source. Then, we computed the relative gain for each detector using the two measurements.

\begin{figure}[t]
    \centering
    \includegraphics{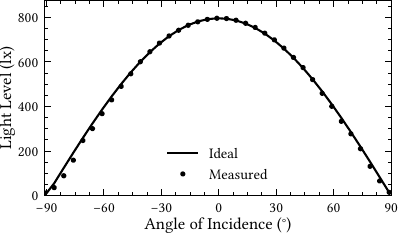}
    \vspace{-0.1in}
    \caption{\textbf{Angular response of the detector.} Across incident angles ranging from $-80\unit{\degree}$ to $+80\unit{\degree}$, the measurements deviate from the ideal cosine response by $3.1\%$ on average.}
    \label{fig:angular-response-supp}
    \vspace{-0.05in}
\end{figure}

In addition, we have measured the angular response of the detector along one dimension by placing the detector on a manual rotation stage. For this experiment, a controlled light source illuminates the bare detector from a distance. \Cref{fig:angular-response-supp} shows the measured light level (in lux) as a function of the angle of incidence of the light source. Over incident angles ranging from $-80\unit{\degree}$ to $+80\unit{\degree}$, the measurement deviates from the ideal cosine response by an average of $3.1\%$.

When the detectors are installed in the prototype, the 3D printed shell slightly occludes their field of view, which in turn impacts their angular response. Using the known geometry of the 3D model, we have calculated that the shell blocks rays with an angle of incidence $\theta \ge 78.5\unit{\degree}$ from reaching the center of each detector. Given that the foreshortened solid angle is small at these large angles of incidence, the occlusion caused by the shell has little impact on the detector's irradiance.

\subsection{Energy Harvesting}
To characterize the energy harvesting performance, we measured the change in energy stored in a capacitor (connected to VSTOR of the BQ25570 energy harvester) without a load attached. The harvested power was computed by dividing the change in stored energy ($\unit{\joule}$) by the time (s). In this case, we operated the sensor at 30 FPS with a short settling time of $50\,\unit{\micro\second}$.

\section{Adaptive Framerate Algorithm}
The power consumption of the irradiance camera is roughly linear with the framerate. Therefore, we could adaptively adjust the camera's framerate so that it only transmits when it has harvested enough energy to do so. This could be implemented as follows: when the irradiance camera wakes up, it reads out all 49 measurements. If the camera has enough stored energy (as measured by the voltage across the supercapacitor), the camera wirelessly transmits those measurements. Then, the camera goes back to sleep for a time that is inversely proportional to the average irradiance measurement over the last 60 seconds. Even if the camera does not have enough energy to wirelessly transmit, this algorithm ensures that it still periodically wakes up to measure irradiance. This allows the camera to adapt to changes in illumination without waiting to harvest enough energy for wireless transmission.

\section{Measured Irradiance Functions in Real Environments}
\begin{figure*}[t]
    \centering
    \includegraphics{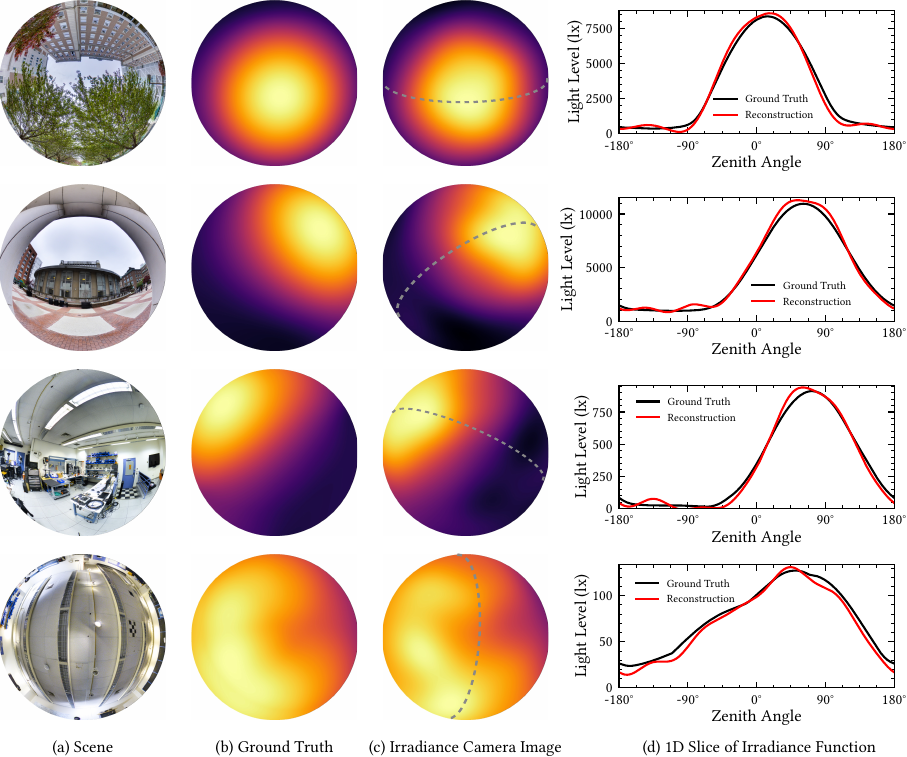}
    \caption{
        \textbf{Irradiance functions measured by the prototype camera in four environments.}
        (a)~HDR image of the scene captured by a Ricoh Theta Z1.
        (b)~Ground truth irradiance function computed using the HDR fisheye image.
        (c)~Irradiance function measured by FluxCam.
        (d)~A 1D slice through the dashed gray lines in~(c).
    }
    \label{fig:real-irradiance-supp}
\end{figure*}

\Cref{fig:real-irradiance-supp} shows the reconstructed irradiance function using measurements made by the prototype camera in four real environments.  We used a Ricoh Theta Z1 to capture an HDR spherical image in each scene, which was used to compute the ground truth irradiance function. To compare the ground truth and measured irradiance functions, we first need to align the poses of the Ricoh Theta Z1 camera and FluxCam. To do this, we begin with a coarse estimate of the relative rotation between the two cameras. Then, we refine this estimate by searching over all rotations within a $20\unit{\degree}$ cone to find the rotation that yields the highest normalized cross-correlation between the ground truth and measured irradiance functions. In this case, we used a long settling time of $20\,\unit{\milli\second}$ to produce high fidelity irradiance measurements.

\section{Estimating Egomotion}

\subsection{Simulation Details}
We used Mitsuba~\cite{wenzeljakobMitsuba3Renderer2022} to simulate the irradiance camera in an urban environment. For each camera location, we rendered the irradiance of its 49 detectors while accounting for their different viewpoints. The irradiance of each detector was simulated using a direct illumination renderer with $16{,}000$ samples. Then, we applied a random rotation and small translation, ranging from $0\,\unit{\meter}$ to $0.1\,\unit{\meter}$, which corresponds to a maximum speed of $2\,\unit{\meter/ \second}$ sampled at 30 FPS, and rendered the measurements in this new pose. In total, this process created $1{,}120{,}250$ samples (pairs of irradiance camera measurements) for training, $160{,}617$ samples for validation, and $319{,}712$ samples for testing.

\subsection{Network Architecture and Training}
The network is a transformer, in which each measurement produced by the irradiance camera corresponds to a single token with an embedding dimension of $D=512$. The inputs to the network are the $49$ measurements (in units of volts) at the two poses, and the output of the network is the rotation matrix between the two camera poses. The network was trained by minimizing the relative angle between the predicted and ground truth rotations using the AdamW optimizer~\cite{loshchilovDecoupledWeightDecay2019} with a learning rate of 1e-4.

\subsection{Results}
\begin{figure}[t]
    \centering
    \includegraphics{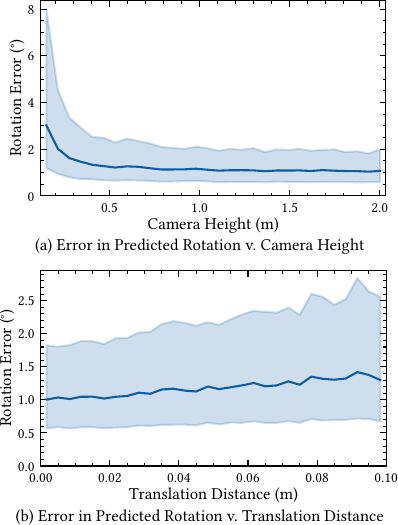}
    \vspace{-0.05in}
    \caption{
        \textbf{Performance of egomotion estimation.}
        (a)~Error in the predicted rotation as a function of camera height above the ground.
        (b)~Error in the predicted rotation as a function of the translation distance. The solid blue line shows the median error, and the light blue region denotes the interquartile range. 
    }
    \label{fig:egomotion-supp}
    \vspace{-0.1in}
\end{figure}

Over the $319{,}712$ test samples, the average error between the predicted and ground truth rotation was $2.25\unit{\degree}$, and the median error was $1.17\unit{\degree}$. \Cref{fig:egomotion-supp} shows the predicted rotation error as a function of the irradiance camera's height above the ground and translation distance. The prediction error is higher when the camera is near the ground or translated by a larger distance. This is expected since the impact of translation on the measured irradiance function is amplified by nearby objects and large translations.

\subsection{Estimating Orientation using Real Measurements}
To compute the FluxCam's orientation from its measurements, we use the trained network to estimate the relative orientation between individual frames (each frame is a set of 49 irradiance measurements). Rather than estimating the relative orientation between consecutive frames, we use the network to estimate the orientation between two frames that are separated in time. We found empirically that this produced better estimates of the camera orientation than using consecutive frames. To combine the estimated orientations, we first scale the angle of each relative rotation to account for the fact that they were computed over a longer time horizon. Then, we accumulate the relative rotations by applying them in sequence. Finally, we filter the resulting camera orientation in quaternion space using a Savitzky--Golay filter.

\section{Estimating Sky Conditions}

\subsection{Experimental Details}
We used FluxCam's measurements to estimate the current sky conditions. To do this, we fit a low-dimensional analytical sky model~\cite{perezAllweatherModelSky1993} to the measurements. 
Due to the spectral response of FluxCam's photovoltaics, the prototype measures illuminance (in lux). Therefore, we first convert FluxCam's measurements to irradiance (in $\unit{\watt\per\meter\squared}$) using a luminous efficacy of $109\,\unit{\lumen\per\watt}$. This luminous efficacy was computed using the global spectrum in the reference air mass 1.5 standard spectra~\cite{am15spectra}. 
To minimize the impact of the ground and nearby buildings on the measured irradiances, we placed FluxCam inside a cylinder with black walls. This cylinder served as an aperture that blocked the irradiance camera's view of the ground and buildings near the horizon. Since the geometry of this shading cylinder is known, we accounted for its impact on each detector's view of the sky. 

For any given sky model, we can apply the aperture caused by the shading cylinder to the corresponding radiance function and then compute what the irradiance measurements would be. Using this approach, we fit the Perez sky model (i.e.,~a model of the sky's radiance function) to the measured irradiances. Additionally, we ignored the measurements produced by the detectors that are oriented sideways or toward the ground.

In order to fit the sky model, we assume that the pose of the irradiance camera in the earth coordinate frame (i.e.,~its orientation with respect to true north) is known. Additionally, we use the time of day and rough GPS location to compute the sun's position in the sky, which is an input to the sky model. 

We have also assumed that the global horizontal irradiance (GHI) is known. This is reasonable since an irradiance camera could be positioned such that one detector has a clear view of the sky, and hence it would directly measure the GHI. In our case, we measured the GHI using a silicon pyranometer (EKO Instruments ML-01). Since GHI is known, the only sky parameter to fit is the diffuse horizontal irradiance (DHI), which specifies the irradiance of a horizontal surface due to the sky, excluding the sun. Note that in the Perez sky model, the direct normal irradiance (DNI) is computed analytically from the GHI and DHI.

\subsection{Validation in Simulation}
We validated the accuracy of this approach for measuring the sky conditions in simulation. We used the Perez sky model to generate the sky's radiance function at every hour of a typical meteorological year in Manhattan. We ignored times when the sun's elevation was lower than $12\unit{\degree}$. Then we simulated the irradiance measurements produced by the camera inside a shading cylinder that blocks objects in the scene with an elevation of $12\unit{\degree}$ or lower. Additionally, this simulation emulates the characteristics of a real sensor that measures irradiance (in $\unit{\watt\per\meter\squared}$) and saturates at 
$1{,}200\,\unit{\watt\per\meter\squared}$ (corresponding to $3.3\,\unit{\volt}$). We assume that the voltage produced by the sensor is proportional to its irradiance, and we model sensor noise by adding $1\,\unit{\milli\volt}$ RMS Gaussian noise to the measurements. We also assume that the shading cylinder is completely black, i.e.,~the radiance from the cylinder is 0. As before, we also assume that the GHI and the camera's pose in the earth coordinate frame are known.

Over 2{,}494 different skies at different times of day throughout the year, this approach estimated the DHI with a root mean square error (RMSE) of $8.3\,\unit{\watt\per\meter\squared}$ and the DNI with an RMSE of $76.0\,\unit{\watt\per\meter\squared}$. The estimated Perez clearness had an RMSE of $0.28$, and the estimated Perez brightness had an RMSE of $0.01$.

\end{document}